\documentclass[runningheads]{llncs}

\usepackage{eccv}

\usepackage{eccvabbrv}

\usepackage{graphicx}
\usepackage{booktabs}
\usepackage{multirow}
\usepackage{pifont}

\usepackage[accsupp]{axessibility}  % Improves PDF readability for those with disabilities.

\usepackage{hyperref}

\usepackage{orcidlink}

\begin{document}
	
% ---------------------------------------------------------------
% REVIEW: Replace with your title
\title{OvDSGG: End-to-End Open-Vocabulary\\
Dynamic Scene Graph Generation}

% REVIEW: If the paper title is too long for the running head, you can set
\titlerunning{OvDSGG: End-to-End Open-Vocabulary Dynamic SGG}

% FINAL: Replace with your author list.
\author{John Helsby\inst{1}\fnmsep\thanks{These authors contribute equally to this work.}\orcidlink{0009-0002-4384-5452} \and
    Yi Yang\inst{2}\fnmsep$^{\star}$\orcidlink{0009-0001-1362-2269} \and
    Bodo Rosenhahn\inst{2}\orcidlink{0000-0003-3861-1424} \and \\
    Michael Ying Yang\inst{1}\fnmsep\thanks{Corresponding author.}\orcidlink{0000-0002-0649-9987}}
% TODO update orcid link

% FINAL: Replace with an abbreviated list of authors.
\authorrunning{J. Helsby et al.}

% FINAL: Replace with your institution list.
\institute{
    University of Bath, UK \and
    Leibniz Universität Hannover, Germany 
}

\maketitle

\begin{abstract}
Dynamic scene graphs (DSGs) capture spatio-temporal interactions across videos as $\langle$subject, predicate, object$\rangle$ triplets, and underpin downstream tasks such as video captioning, video question answering, and action analysis. 
However, end-to-end dynamic scene graph generation (DSGG) methods are closed-set: they recognize only objects and predicates from a fixed training vocabulary and struggle with the long-tailed distribution of rare concepts, severely limiting their real-world applicability. 
Existing open-vocabulary models typically inherit pretrained large language models, resulting in multi-stage training and inference with substantial cost. 
We introduce OvDSGG, the first end-to-end framework for open-vocabulary DSGG. OvDSGG builds on top of an open-vocabulary Spatial Backbone and a Temporal Backbone; we further propose a Triplet Feature Extraction Module that bridges them, and a Visual-Language Alignment Module that preserves open-vocabulary recognition by learning an adaptive decision boundary in the joint visual-language feature space, without expensive knowledge distillation in existing methods. We further introduce a rigorous open-vocabulary DSGG benchmark adapted from Action Genome, with disjoint Base/Novel splits for both objects and predicates. OvDSGG significantly outperforms open-vocabulary baselines across all metrics, with zero-shot Recall@$K$ scores 10.0--20.4 percentage point higher than the next-best baseline, while on closed-set DSGG remaining competitive with state-of-the-art models. 
Code and benchmark are publicly available at \url{https://github.com/jhelsby/OvDSGG/}.
\end{abstract}

%-------------------------------------------------------------------------
\section{Introduction}
\label{sec:intro}

Scene graphs~\cite{johnson2015image} encode visual scenes in images as a structured set of triplets: $\langle\textit{subject}, \textit{predicate}, \textit{object}\rangle$. Dynamic Scene Graphs (DSGs) extend this representation to video by adding a temporal dimension, capturing how interactions evolve over time~\cite{ji2019actiongenomeactionscomposition}. The accompanying task of Dynamic Scene Graph Generation (DSGG) supports downstream applications such as video captioning, video question answering, and action analysis~\cite{nguyen2025hyperglmhypergraphvideoscene}.

\begin{figure}[t]
    \centering
    \includegraphics[width=0.85\linewidth]{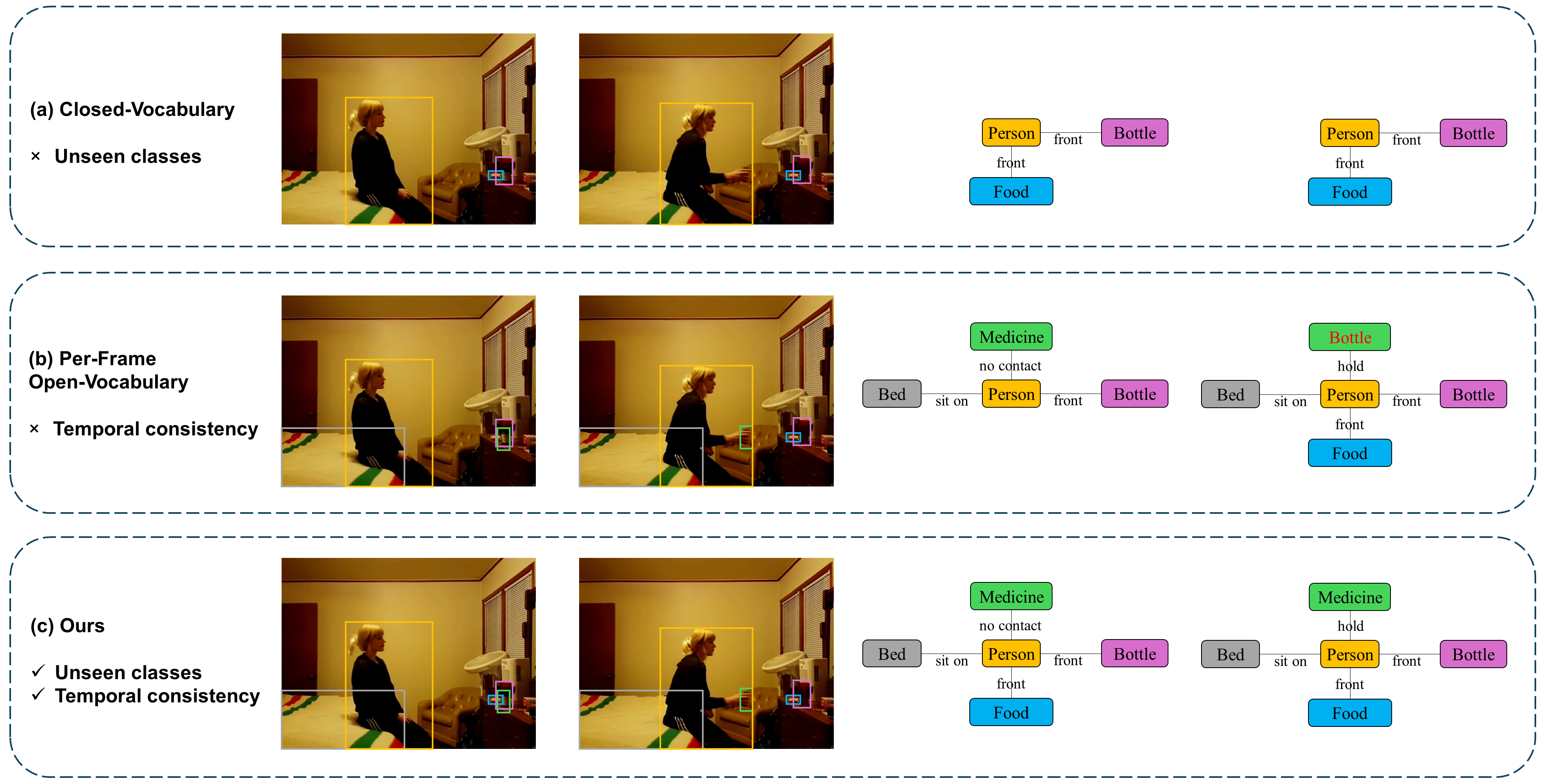}
    \caption{Comparison between existing DSGG methods and our OvDSGG. (a) Closed-vocabulary DSGG fails to detect objects and predicates that are unseen in training data. (b) Per-frame open-vocabulary models do not capture temporal dynamics. (c) Our OvDSGG is an end-to-end, open-vocabulary model for DSGG. }
    \label{fig:teaser}
    \vspace{-4mm}
\end{figure}

Despite rapid progress in DSGG, existing end-to-end methods share a critical limitation: they are \emph{closed-set}, capable only of identifying objects and predicates from a predefined vocabulary, as shown in Figure~\ref{fig:teaser} (a). They also struggle with the long-tailed distribution of rarely seen concepts~\cite{zhu2022scenegraphgenerationcomprehensive,chen2025datamodelingfullyopenvocabulary}. This restricts their use in real-world scenarios, where novel objects and relationships frequently emerge.

For static scene graph generation (SGG), this restriction has been substantially relaxed by recent work on \emph{open-vocabulary} scene graph generation~\cite{he2022openvocabularyscenegraphgeneration,zhang2023openvocabularyscenegraphgeneration,chen2024expanding,chen2025datamodelingfullyopenvocabulary}, which leverages pretrained vision--language models to recognize unseen objects and unseen relationships in images. 
While they can generate per-frame open-vocabulary scene graphs, they lack the capability to capture the temporal dynamics and consistency in videos, as illustrated in Figure~\ref{fig:teaser} (b). 

We study this setting with OvDSGG, a novel end-to-end framework that is fully open-vocabulary in both objects and predicates, as illustrated in Figure~\ref{fig:teaser}~(c). OvDSGG integrates the open-vocabulary image SGG framework OvSGTR~\cite{chen2024expanding,chen2025datamodelingfullyopenvocabulary} with the one-stage DSGG model OED~\cite{wang2024oedonestageendtoenddynamic}: it retains OvSGTR's open-vocabulary spatial backbone, adopts OED's temporal context aggregation module, and connects the two via two novel components: a Triplet Feature Extraction Module and a Visual--Language Alignment Module.

Two obstacles make this adaptation non-trivial. (i) The temporal module requires a compact per-query triplet representation, but Grounding DINO emits only per-object queries; our Triplet Feature Extraction Module (Section~\ref{sec:triplet}) lifts these queries into paired instance and predicate features and concatenates them into the triplet representation the temporal module consumes. (ii) Open-vocabulary recognition must be preserved during temporal training, yet OvSGTR's knowledge-distillation retention loss requires a teacher forward pass for every frame of every training window, and is thus prohibitively expensive on video; our Visual--Language Alignment Module (Section~\ref{sec:align_module}) instead decouples the semantic prior from the decision boundary by using text embeddings to initialize learnable classification heads, thus retaining novel-class prototypes structurally and dispensing with the distillation loss altogether. (iii) No open-vocabulary benchmark exists for DSGG; we provide one by partitioning Action Genome into disjoint Base/Novel splits compatible with the OvSGTR pretraining used to initialize OvDSGG.

In summary, the contributions of this paper are:
\begin{enumerate}
    \item \textbf{OvDSGG}, the first end-to-end open-vocabulary DSGG framework, capable of predicting both unseen objects and unseen predicates from video. We propose a Triplet Feature Extraction Module that facilitates temporal aggregation, and a Visual--Language Alignment Module that retains open-vocabulary capacity without the need of expensive knowledge distillation.
    \item A formal \textbf{open-vocabulary DSGG benchmark} on Action Genome~\cite{ji2019actiongenomeactionscomposition}, with disjoint Base/Novel splits enabling evaluation across seen and unseen object and relation categories.
    \item \textbf{Experimental results} demonstrating that OvDSGG outperforms all baselines in the open-vocabulary setting, with zR@$K$ scores 10.0--20.4 percentage point higher than the next-best baseline, while remaining competitive with state-of-the-art methods in the closed-set setting.
\end{enumerate}

%-------------------------------------------------------------------------
\section{Related Work}
\label{sec:related}
\textbf{Scene graph generation.} Following Johnson \etal~\cite{johnson2015image}, scene graphs have become a standard structured representation for images, with Visual Genome~\cite{krishna2017visual} as the canonical benchmark. Early SGG methods are two-stage, first detecting objects via Faster-RCNN~\cite{ren2015faster} and then classifying relations between proposals~\cite{lu2016visual,zhu2022scenegraphgenerationcomprehensive}. 
To avoid the quadratic relation-classification cost and the difficulty of joint optimization, one-stage end-to-end methods built on DETR~\cite{carion2020end} have emerged, including SGTR~\cite{li2022sgtr}, RelTR~\cite{cong2023reltr}, and EGTR~\cite{im2024egtr}. 
A persistent obstacle across both paradigms is the long-tailed distribution of relationship annotations: a few frequent geometric and possessive predicates dominate while semantically rich predicates are rare~\cite{zellers2018neural,tang2020unbiased}. This biases models toward frequent predicates and degrading generalization to rare triplets. A line of \emph{unbiased} SGG work counteracts this, \eg\ via causal interventions that remove the dataset frequency bias at inference~\cite{tang2020unbiased}. Open-vocabulary SGG offers a distinct but equally effective approach: it replaces closed-set classifiers with vision--language alignment, so that rare and unseen concepts can be recognized directly.

\noindent\textbf{Dynamic scene graph generation.} The DSGG task and the Action Genome dataset were introduced by Ji \etal~\cite{ji2019actiongenomeactionscomposition}, providing 10K videos with frame-level triplet annotations. Early DSGG methods adopt the two-stage paradigm, with separate object detection and relation classification stages~\cite{li2022anticipatory,feng2023exploiting}. STTran~\cite{cong2021spatial} introduced a spatio-temporal transformer combining a Faster-RCNN backbone with spatial and temporal decoders, while TPT~\cite{zhang2024e2evideo} was the first transformer-based end-to-end DSGG method, though it remained two-stage. Most recently, OED~\cite{wang2024oedonestageendtoenddynamic} proposed an one-stage, end-to-end architecture that extends DETR across both spatial and temporal dimensions, achieving state-of-the-art results on Action Genome. However, like all prior DSGG methods, OED is restricted to a closed-set vocabulary and inherits the long-tail bias of its training data. TEMPURA~\cite{nag2023unbiased} targets this bias directly, combining temporal consistency modeling with uncertainty-guided debiasing, but it remains a two-stage, closed-set model and is outperformed by the end-to-end OED.

\noindent\textbf{Open-vocabulary scene graph generation.} Open-vocabulary object detectors~\cite{du2022learning,li2022grounded,wu2023aligning,zareian2021open,zhong2022regionclip} leverage pretrained vision--language models (VLMs), often building on CLIP~\cite{radford2021learning}. GLIP~\cite{li2022grounded} reformulates detection as image--text matching, and Grounding DINO~\cite{liu2024grounding} combines this with the DETR-like detector DINO~\cite{zhang2022dino} to yield a strong open-vocabulary detector. Building on these advances, He \etal~\cite{he2022openvocabularyscenegraphgeneration} and Zhang \etal~\cite{zhang2023openvocabularyscenegraphgeneration} introduced open-vocabulary SGG models capable of recognizing unseen objects, and the more recent OvSGTR framework~\cite{chen2024expanding,chen2025datamodelingfullyopenvocabulary} achieved state-of-the-art results on Visual Genome by extending open-vocabulary recognition to predicates as well. OvSGTR combines a Swin Transformer~\cite{liu2021swin} visual encoder, a BERT~\cite{devlin2019bert} text encoder, and a DETR-style relational transformer in a single end-to-end network. A parallel line of work instead builds open-vocabulary scene graph generators on large pretrained VLMs and multimodal language models: PGSG~\cite{li2024pixels} recasts SGG as image-to-text generation with a generative VLM; Robin~\cite{park2025synthetic} instruction-tunes a multimodal language model to produce dense scene graphs; and DovSG~\cite{yan2025dynamic} assembles open-vocabulary 3D scene graphs from a pipeline of foundation models. These methods inherit the broad semantic coverage of their pretrained backbones and report strong open-vocabulary performance, but they are multi-stage systems with substantial training and inference cost, which therefore do not fit into an end-to-end video DSGG framework.

\noindent\textbf{Research gap.} While open-vocabulary SGG has matured on static images, the corresponding video task remains less explored. State-of-the-art DSGG models such as OED are closed-set and suffer from the same long-tail bias that open-vocabulary methods mitigate for images. We note that OED and OvSGTR share a common DETR lineage, which makes them amenable to re-adaptation and integration. This motivates OvDSGG, the one-stage, end-to-end, fully open-vocabulary DSGG framework presented next.

%-------------------------------------------------------------------------
\section{Method}
\label{sec:method}

%-------------------------------------------------------------------------
\subsection{Problem Formulation}
\label{sec:problem}

A dynamic scene graph is a structured representation, capturing evolving entities and visual interactions within a video. Nodes correspond to localized visual entities, while directed edges denote pairwise interactions.

Given an input video $\{V_t\}_{t=1}^{H}$ of $H$ sequential frames, the $k$-th entity node $o_t(k)$ in frame $t$ is parameterized by its semantic category $c_t(k) \in \mathcal{C}$ and its normalized bounding-box coordinates $b_t(k) \in [0, 1]^4$, where $\mathcal{C}$ defines the vocabulary of all object classes. Directed edges capture interactions from a source \textit{subject} node ($i$) to a destination \textit{object} node ($j$), characterized by a predicate $p_t (i, j) \in \mathcal{P}$ from the predicate vocabulary $\mathcal{P}$. There can be multiple predicates between a single subject--object pair. A complete visual relationship triplet is
\begin{equation}
    r_t (i, j) = (o_t(i),\, p_t (i, j),\, o_t(j)),
\end{equation}
and the scene graph for frame $t$ is the set $G_t = \{r_t^k\}_{k=1}^{K}$ of all valid triplets.

We define two task variants:

\noindent\textbf{Closed-Set DSGG} (\texttt{CS-DSGG}). Generate $\{\hat{G}_t\}_{t=1}^{H}$ with $\mathcal{C}_{\text{train}} = \mathcal{C}_{\text{eval}}$ and $\mathcal{P}_{\text{train}} = \mathcal{P}_{\text{eval}}$.

\noindent\textbf{Open-Vocabulary DSGG} (\texttt{OV-DSGG}). Vocabularies partition into disjoint Base and Novel subsets, $\mathcal{C} = \mathcal{C}_{\text{base}} \cup \mathcal{C}_{\text{novel}}$ and $\mathcal{P} = \mathcal{P}_{\text{base}} \cup \mathcal{P}_{\text{novel}}$. Training accesses only Base annotations; evaluation uses the full $\mathcal{C}$ and $\mathcal{P}$.

%-------------------------------------------------------------------------
\subsection{Overview}
\label{sec:overview}

OvDSGG is an end-to-end DSGG framework that comprises the following four components.
The Spatial Backbone (Section~\ref{sec:gdino}) generates per-query entity features from Grounding DINO. 
The Triplet Feature Extraction Module (Section~\ref{sec:triplet}) projects these entity queries into paired instance and predicate features at the triplet level.
The Temporal Backbone (Section~\ref{sec:temporal}) aggregates cross-frame context via OED's Progressively Refined Module.
Finally, the Visual--Language Alignment Module (Section~\ref{sec:align_module}) instantiates open-vocabulary classifiers via BERT-initialized linear heads and produces the final triplet predictions.

\begin{figure}[htbp]
    \centering
    \includegraphics[width=0.9\linewidth]{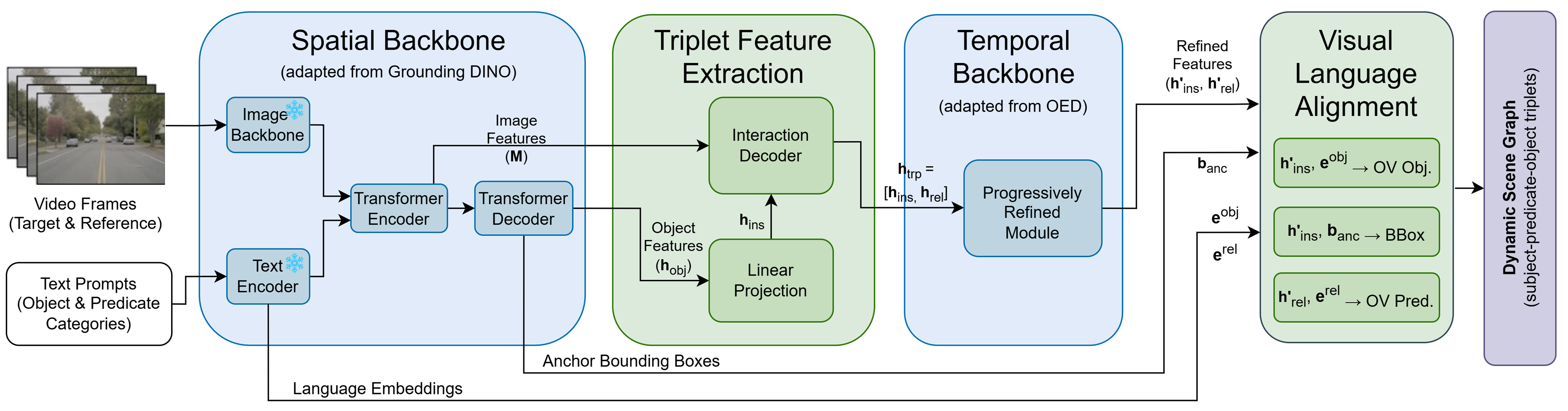}
    \caption{Overview of our OvDSGG architecture. The Spatial Backbone extracts per-frame visual and language features, which are processed by Triplet Feature Extraction to form triplet representations. The Temporal Backbone further aggregates temporal information from reference frames. The Visual--Language Alignment Module uses temporally enhanced triplet features and language embeddings to produce open-vocabulary classification results.}
    \vspace{-8mm}
    \label{fig:ovdsgg_architecture}
\end{figure}

%-------------------------------------------------------------------------
\subsection{Spatial Backbone with Grounding DINO}
\label{sec:gdino}

OvDSGG employs Grounding DINO~\cite{liu2024grounding} as its foundational vision--language detector, extracting entity queries and  visual features from each input frame. 

Following the prompting strategy of OvSGTR~\cite{chen2025datamodelingfullyopenvocabulary}, we format object classes into a single unified prompt, concatenating class names separated by periods (e.g., ``\texttt{person\,.\,bag\,.\,bed\,.\,}\dots''). The prompt is  encoded by BERT~\cite{devlin2019bert}, yielding per-class semantic embeddings $\mathbf{e}^{\text{obj}}_c \in \mathbb{R}^d$ for each $c \in \mathcal{C}$. These embeddings guide the vision backbone via cross-attention inside Grounding DINO. Separately, the predicate class names are encoded by the same language backbone, yielding embeddings $\mathbf{e}^{\text{rel}}_p \in \mathbb{R}^d$ for each $p \in \mathcal{P}$. The predicate embeddings are not used for visual grounding but are stored for downstream initialization of the predicate classification heads (Section~\ref{sec:align_module}).

For each input frame, Grounding DINO outputs a fixed set of $N_q$ entity queries. We extract their content representations $h_{\text{obj}} \in \mathbb{R}^{N_q \times d}$ and the corresponding anchor bounding boxes $b_{\text{anc}} \in [0,1]^{N_q \times 4}$. $h_{\text{obj}}$ is already cross-modally enhanced and carries open-vocabulary semantics, as a result of Grounding DINO's vision--language fusion. 
% This property is later exploited by the Visual--Language Alignment Module (Section~\ref{sec:align_module}) for open-vocabulary classification. 
The detector also outputs dense memory feature maps $M \in \mathbb{R}^{N_v \times d}$ (with $N_v$ visual tokens) from its multi-scale encoder. 

In summary, Grounding DINO outputs $\langle h_{\text{obj}},\, b_{\text{anc}},\, M,\,\mathbf{e}^{\text{obj}}_c,\,\mathbf{e}^{\text{rel}}_p\rangle$, a per-frame interface between the Spatial Backbone and the downstream modules.

%-------------------------------------------------------------------------
\subsection{Triplet Feature Extraction Module}
\label{sec:triplet}

The entity queries $h_{\text{obj}}$ from Grounding DINO encode each detected object in isolation and are not directly amenable to relational reasoning. We therefore decouple instance and predicate representations into separate branches.

\noindent\textbf{Instance features.} A learnable linear projection lifts the object-centric query representations to instance features:
\begin{equation}
    h_{\text{ins}} = \mathbf{W}_{\text{ins}}\, h_{\text{obj}} \in \mathbb{R}^{N_q \times d}\,,
\end{equation}
with $\mathbf{W}_{\text{ins}} \in \mathbb{R}^{d \times d}$ applied to each query.

\noindent\textbf{Predicate features.} An interaction decoder produces predicate features by attending the instance queries against the dense memory feature maps $M$, gathering the contextual visual evidence needed for relational reasoning:
\begin{equation}
    h_{\text{rel}} = \text{InteractionDecoder}(h_{\text{ins}};\, M) \in \mathbb{R}^{N_q \times d}\,.
\end{equation}
The decoder consists of three Transformer decoder layers, each comprising self-attention over the queries, cross-attention to $M$, and a feed-forward block.

\noindent\textbf{Triplet representation.} The per-query triplet representation is the channel-wise concatenation of the instance and predicate features:
\begin{equation}
    h_{\text{trp}} = [h_{\text{ins}};\, h_{\text{rel}}] \in \mathbb{R}^{N_q \times 2d}\,.
\end{equation}
This representation jointly carries the spatial and relational information for each query, and serves as input to the Temporal Backbone (Sec.~\ref{sec:temporal}) and the classification heads (Sec.~\ref{sec:align_module}).

%-------------------------------------------------------------------------
\subsection{Temporal Backbone}
\label{sec:temporal}

For temporal reasoning, we adopt OED's Progressively Refined Module~\cite{wang2024oedonestageendtoenddynamic}. This module operates on the per-query triplet representation $h_{\text{trp}}$.

Given the triplet representations for a target frame ($h^{\text{tgt}}_{\text{trp}}$) and its $n$ reference frames ($h^{\text{ref}}_{\text{trp}}$), we rank the reference queries by a per-query confidence score
\begin{equation}
    p = p_{\text{obj}} \cdot p_{\text{attn}} \cdot p_{\text{spat}} \cdot p_{\text{cont}}\,,
\end{equation}
where $p_{\text{obj}}$ is the maximum-class probability from the object classifier of Section~\ref{sec:align_module}, restricted to foreground classes; and $p_{\text{attn}}$, $p_{\text{spat}}$, $p_{\text{cont}}$ are the maximum-class probabilities from the three predicate-group classifiers (also Section~\ref{sec:align_module}). 

The Progressively Refined Module then distills temporal context via a cascade of $m$ layers. In each layer $i \in \{1, \dots, m\}$, we retain the top-$k_i$ most confident reference queries and cross-attend the target into them:
\begin{equation}
\begin{gathered}
    h^{\text{ref};k_i}_{\text{trp}} = \text{Top-}K\!\left(h^{\text{ref}}_{\text{trp}},\, k_i\right), \\
    h^{\text{tgt}}_{i} = \text{FFN}\!\left(\text{CrossAttn}\!\left(\text{SelfAttn}(h^{\text{tgt}}_{i-1}),\, h^{\text{ref};k_i}_{\text{trp}}\right)\right),
\end{gathered}
\end{equation}
with $h^{\text{tgt}}_{0} = h^{\text{tgt}}_{\text{trp}}$. The selection threshold $k_i$ is progressively reduced across layers (e.g., $80n \to 50n \to 30n$), distilling temporal context while filtering background noise. After refinement, $h^{\text{tgt}}_{m}$ is split along the channel dimension into temporally enriched instance ($h'_{\text{ins}}$) and relation ($h'_{\text{rel}}$) features, which feed the final classification heads of Section~\ref{sec:align_module}.

%-------------------------------------------------------------------------
\subsection{Visual--Language Alignment Module}
\label{sec:align_module}

The Visual--Language Alignment Module produces the final open-vocabulary triplet predictions. It instantiates a set of classification heads whose weights are initialized from the BERT text embeddings $\mathbf{e}^{\text{obj}}_c$ and $\mathbf{e}^{\text{rel}}_p$ from the Spatial Backbone (Section~\ref{sec:gdino}), aligning visual predictions to the language semantic space, so that categories unseen during training can still be recognized. The heads are applied to the temporally enriched features from the Temporal Backbone (Section~\ref{sec:temporal}) to produce frame-level predictions.

Unlike OvSGTR which adopts a cosine similarity loss between the frozen BERT text embeddings and the output features, 
OvDSGG decouples semantic prior and decision boundary. Text embeddings are used to initialize the weights of learnable linear heads, supplying the open-vocabulary prior, while admitting per-class capacity to adapt base-class boundaries to the target domain. Novel categories are excluded from the object prompt, masked from the predicate loss (Section~\ref{sec:training}), and receive no gradient. The rows of the head matrices corresponding to novel categories remain at their text-embedding initialization throughout training. 

\noindent\textbf{Object classification.} 
A linear head maps instance features to per-query object probabilities:
\begin{equation}
    \hat{p}^{\text{obj}}_q = \sigma\!\left(\mathbf{W}^{\text{obj}}\, h^{q}_{\text{ins}}\right) \in [0,1]^{|\mathcal{C}|}\,,
\end{equation}
where $\sigma(\cdot)$ is the element-wise sigmoid. Writing $\tilde{\mathbf{e}} = \mathbf{e}/\|\mathbf{e}\|_2$ for $\ell_2$-normalization, the weight matrix $\mathbf{W}^{\text{obj}} \in \mathbb{R}^{|\mathcal{C}| \times d}$ is initialized row-wise from the object-class text embeddings:
\begin{equation}
    \mathbf{W}^{\text{obj}}_{c,:} = \tilde{\mathbf{e}}^{\text{obj}}_c, \quad c \in \mathcal{C}\,.
\end{equation}

\noindent\textbf{Bounding-box regression.} Two three-layer MLPs respectively predict subject and object coordinate offsets from the instance features:
\begin{equation}
    \Delta b^{l}_q = \text{MLP}^{l}(h^{q}_{\text{ins}}), \quad l \in \{\text{sub},\, \text{obj}\}\,.
\end{equation}
For gradient stability, offsets are added in inverse-sigmoid space to the anchor boxes $b_{\text{anc}}$:
\begin{equation}
\label{eqn:bbox_update}
    \hat{b}^{l}_q = \sigma\!\left(\sigma^{-1}(b_{\text{anc},q}) + \Delta b^{l}_q\right)\,.
\end{equation}

\noindent\textbf{Predicate classification.} Reflecting Action Genome's predicate structure, we instantiate three independent linear heads, one per group $g \in \mathcal{R} = \{\text{attn},\, \text{spat},\, \text{cont}\}$:
\begin{equation}
    \hat{p}^{g}_q = \sigma\!\left(\mathbf{W}^{g}\, h^{q}_{\text{rel}}\right) \in [0,1]^{|\mathcal{P}^g|}\,, \quad g \in \mathcal{R}\,,
\end{equation}
where $\mathcal{P}^k$ is the predicate vocabulary of group $g$, and each weight matrix is initialized analogously from the predicate text embeddings:
\begin{equation}
    \mathbf{W}^{g}_{p,:} = \tilde{\mathbf{e}}^{\text{rel}}_p, \quad p \in \mathcal{P}^g\,.
\end{equation}

% The composition of the vision-guiding object prompt depends on the evaluation setting. 
For \texttt{CS-DSGG}, the prompt encompasses the full $\mathcal{C}$. % and remains static during training and evaluation. 
For \texttt{OV-DSGG}, the training prompt is restricted to $\mathcal{C}_{\text{base}}$ and expanded to $\mathcal{C}$ at evaluation. For predicates, the heads always span the full $\mathcal{P}$; open-vocabulary transfer is instead enforced via a gradient mask on the loss (Section~\ref{sec:training}).

At inference, the classification heads operate on the temporally refined features $h'_{\text{ins}}$ and $h'_{\text{rel}}$ from the Temporal Backbone (Section~\ref{sec:temporal}).

%-------------------------------------------------------------------------
\subsection{Training and Inference}
\label{sec:training}

\noindent\textbf{Training.} For each target frame, the model predicts a fixed set of $N_q$ candidate triplets $T = \{t_i\}_{i=1}^{N_q}$. We use Hungarian Matching~\cite{carion2020end} to find the optimal bipartite matching $\hat{\pi}$ between $T$ and the ground-truth set $G = \{r_i\}_{i=1}^{N_q}$ (padded with $\varnothing$):
\begin{equation}
    \hat{\pi} = \underset{\pi \in \mathfrak{S}_{N_q}}{\arg\min} \sum_{i=1}^{N_q} \mathcal{L}_{\text{match}}(r_i,\, t_{\pi(i)})\,.
\end{equation}
The overall training objective combines classification and bounding-box losses:
\begin{equation}
    \mathcal{L}_{\text{match}} = \mathcal{L}_{\text{obj\_cls}} + \sum_{g \in \mathcal{R}} \mathcal{L}^{g}_{\text{rel\_cls}} + \sum_{l \in \{\text{sub},\, \text{obj}\}} \mathcal{L}^{l}_{\text{box}}\,.
\end{equation}

\noindent\textit{Classification losses.} We use multi-label sigmoid Focal Loss~\cite{lin2017focal} for classification. 
To support open-vocabulary learning, we apply two complementary masking strategies. For object classification, novel categories are excluded from the training object prompt so the detector emits no logits for them; the loss is summed only over base classes:
\begin{equation}
    \mathcal{L}_{\text{obj\_cls}} = \frac{1}{N_{\text{pos}}} \sum_{q=1}^{N_q} \sum_{c \in \mathcal{C}_{\text{base}}} \text{FocalLoss}\!\left(y^{\text{obj}}_{q,c},\, \hat{p}^{\text{obj}}_{q,c}\right)\,,
\end{equation}
where $y^{\text{obj}}_{q,c} \in \{0,1\}$ is the object target for query $q$ at class $c$, and $N_{\text{pos}}$ is the number of annotated positive matches. 
For predicate classification, the heads are structurally fixed to the full vocabulary, so we instead introduce a binary mask $m^{g}_{p} = \mathbf{1} [p \in \mathcal{P}^g_{\text{base}}]$ that restricts the loss to base predicates:
\begin{equation}
    \mathcal{L}^{g}_{\text{rel\_cls}} = \frac{1}{N_{\text{pos}}} \sum_{q=1}^{N_q} \sum_{p \in \mathcal{P}^g} m^{g}_{p} \cdot \text{FocalLoss}\!\left(y^{g}_{q,p},\, \hat{p}^{g}_{q,p}\right)\,.
\end{equation}
These schemes preserve novel categories' text-embedding in the classifier (Section~\ref{sec:align_module}). At evaluation, both sums extend over the full vocabularies $\mathcal{C}$ and $\mathcal{P}^g$.

\noindent\textit{Bounding-box loss.} We use a weighted combination of $\ell_1$ and GIoU~\cite{rezatofighi2019generalized} losses:
\begin{equation}
    \mathcal{L}^{l}_{\text{box}} = \alpha\,\mathcal{L}^{l}_{\ell_1} + \beta\,\mathcal{L}^{l}_{\text{GIoU}}\,,
\end{equation}
with weighting coefficients $\alpha, \beta \in \mathbb{R}^{+}$. Boxes are regressed as offsets to the detector's anchor boxes $b_{\text{anc}}$ (Eq.~\ref{eqn:bbox_update}) rather than from scratch.

\noindent\textit{Partial freezing.} To balance preservation of pretrained open-vocabulary knowledge against temporal adaptation, we adopt the partial-freezing strategy: the visual backbone and BERT are entirely frozen, while the final six layers of Grounding DINO's cross-modal encoder and decoder are fine-tuned.

\noindent\textbf{Inference.} OvDSGG observes features pooled across $n$ reference frames and emits $N_q$ candidate triplets per target frame. The candidates are first pruned by a global object-confidence threshold to retain the top-$K$ queries, then de-duplicated by batched Non-Maximum Suppression over the projected object boxes and category labels. Sliding the temporal window over the video yields the complete DSG sequence.

%-------------------------------------------------------------------------
\section{Experiments}
\label{sec:experiments}

%-------------------------------------------------------------------------
\subsection{Dataset and Open-Vocabulary Splits}
\label{sec:dataset}

We evaluate on Action Genome (AG)~\cite{ji2019actiongenomeactionscomposition}, which provides frame-level scene graph annotations across 234{,}253 frames, with 36 object classes and 26 predicate classes. For \texttt{CS-DSGG} we use the standard AG training/test split.

For \texttt{OV-DSGG}, following OvSGTR~\cite{chen2025datamodelingfullyopenvocabulary}, we partition both object and predicate categories into a 70\% Base / 30\% Novel split. Because OvDSGG is initialized from an OvSGTR checkpoint pretrained on Visual Genome (VG)~\cite{krishna2017visual}, which has only partial overlap with AG, we map categories between the two datasets to prevent data contamination. We (i) inherit the Base/Novel assignments from the VG splits used by the OvSGTR checkpoint (44\% of AG objects and 36\% of AG predicates inherit Base; 11\% and 8\% inherit Novel); (ii) allocate the remaining 44\% of objects and 54\% of predicates absent from VG into Base or Novel by frequency-balanced sampling to reach the 70/30 target; and (iii) resolve ambiguous grouped labels (\eg\ \texttt{cup/glass/bottle}) conservatively, assigning all such grouped categories to Base to avoid contaminating the Novel set. 
Detailed statistics are provided in Supplementary Section~A.

Following standard SGG and DSGG evaluation metrics, we report Recall@$K$ (R@$K$)~\cite{lu2016visual} for overall performance, mean Recall@$K$ (mR@$K$)~\cite{chen2019knowledge} for long-tailed performance, and zero-shot Recall@$K$ (zR@$K$)~\cite{lu2016visual,tang2020unbiased} for unseen triplets.

%-------------------------------------------------------------------------
\subsection{Implementation and Settings}
\label{sec:settings}

The spatial detector uses a Swin-T~\cite{liu2021swin} backbone with $N_q = 900$ entity queries and hidden dimension 256, following OvSGTR. The temporal module uses 3 interaction decoder layers, 3 temporal decoder layers, and a window of 3 reference frames. The \texttt{CS-DSGG} spatial module is initialized from a closed-set OvSGTR checkpoint pretrained on VG; \texttt{OV-DSGG} uses its open-vocabulary equivalent. 
% Experiments use a single NVIDIA RTX A5000 (24\,GB).

Training follows OED's two-stage scheme: spatial module is trained first, then temporal module is fine-tuned. For \texttt{CS-DSGG} the spatial module converged after 3 epochs and the temporal module after 2. 
For \texttt{OV-DSGG} we found that directly fine-tuning the full model during temporal training degraded spatial performance, consistent with the catastrophic-forgetting behavior reported for OvSGTR~\cite{chen2025datamodelingfullyopenvocabulary}. We therefore freeze Grounding DINO and the spatial DSGG components during open-vocabulary temporal training. Our \texttt{OV-DSGG} temporal model was trained for 2 epochs with AdamW at LR $5\!\times\!10^{-5}$, followed by 1 epoch at $1\!\times\!10^{-5}$.

We use \texttt{SGDET} as our evaluation protocol, which requires prediction of subject and object boxes, their classes, and the  predicate, given only the video frames. We consider a predicted box correct if it overlaps with ground truth with IoU $\ge 0.5$. All metrics are reported under the standard \emph{With Constraint} (one predicate per pair) and \emph{No Constraint} (multiple predicates per pair) settings~\cite{cong2021spatial,zhu2022scenegraphgenerationcomprehensive}.

%-------------------------------------------------------------------------
\subsection{Open-Vocabulary DSGG}
\label{sec:results-ov}

Table~\ref{tab:ov-dsgg} compares OvDSGG against three baseline variants we construct, since no prior end-to-end open-vocabulary DSGG model exists to compare against directly: an open-vocabulary adaptation of OED in both its spatial-only and full temporal forms, and OvSGTR as a spatial baseline on AG. 

We report both the spatial-only and full temporal variants of OvDSGG. OvDSGG provides superior scores across all metrics, outperforming the next-best baseline on R@$K$ by 9.6--16.6\,percentage point (pp) and on mR@$K$ by 1.3--8.1\,pp. Crucially, zR@$K$ is 10.0--20.4\,pp higher than the next-best baseline (\eg\ 13.6 vs.\ 3.6 on zR@10). These results support our central claim: by combining Grounding DINO's open-vocabulary object detection with BERT-initialized predicate heads, OvDSGG performs effective open-vocabulary DSGG on both objects and predicates.

\begin{table}[htbp]
  \centering
  \caption{Comparison with baseline methods on Action Genome for Scene Graph Detection (\texttt{SGDET}), in the open-vocabulary setting. Best and second-best are \textbf{bold} and \underline{underlined}, respectively.}
  \resizebox{\linewidth}{!}{%
     \begin{tabular}{l|cccccc|cccccc|cccccc}
    \hline
    \multirow{2}{*}{\texttt{SGDET}, IoU\,$\ge 0.5$} & \multicolumn{6}{c|}{R@$K$ $\uparrow$} & \multicolumn{6}{c|}{mR@$K$ $\uparrow$} & \multicolumn{6}{c}{zR@$K$ $\uparrow$} \\
    % \cline{2-19}
          & \multicolumn{3}{c}{With Constr.} & \multicolumn{3}{c|}{No Constr.} & \multicolumn{3}{c}{With Constr.} & \multicolumn{3}{c|}{No Constr.} & \multicolumn{3}{c}{With Constr.} & \multicolumn{3}{c}{No Constr.} \\
    % \hline
    Method & 10 & 20 & 50 & 10 & 20 & 50 & 10 & 20 & 50 & 10 & 20 & 50 & 10 & 20 & 50 & 10 & 20 & 50 \\
    \hline
    Ov-OED (spatial) \cite{wang2024oedonestageendtoenddynamic} & 22.9 & 26.4 & 29.6 & 21.5 & 26.7 & 34.6 & {11.5} & {13.6} & 15.2 & {14.5} & 21.4 & 32.3 & 2.0 & 2.8 & 3.9 & 3.6 & 8.3 & 13.8 \\
    Ov-OED (temporal) & {23.7} & {27.8} & {31.5} & {21.9} & {27.5} & {36.0} & 11.2 & 13.5 & {15.9} & {14.5} & {22.0} & {33.5} & 0.5 & 0.7 & 1.1 & {3.9} & {8.6} & {14.9} \\
    OvSGTR \cite{chen2024expanding,chen2025datamodelingfullyopenvocabulary} & 12.9 & 16.4 & 20.8 & 12.2 & 16.8 & 23.1 & 7.9 & 9.8 & 11.9 & 9.8 & 14.2 & 20.0 & {3.6} & {4.8} & {6.2} & 3.4 & 5.1 & 8.7 \\
    \hline
    OvDSGG (spatial) & \underline{32.0} & \underline{37.6} & \underline{43.6} & \underline{32.9} & \underline{40.5} & \underline{50.0} & \underline{14.1} & \underline{16.6} & \underline{19.1} & \underline{19.0} & \underline{25.7} & \textbf{36.5} & \textbf{15.5} & \textbf{20.4} & \underline{25.6} & \textbf{17.7} & \textbf{24.6} & \textbf{33.1} \\
    OvDSGG (temporal) & \textbf{33.3} & \textbf{40.1} & \textbf{48.1} & \textbf{33.9} & \textbf{41.5} & \textbf{51.0} & \textbf{16.5} & \textbf{19.8} & \textbf{24.0} & \textbf{19.9} & \textbf{25.8} & \underline{34.8} & \underline{13.6} & \underline{18.8} & \textbf{26.6} & \underline{15.1} & \underline{21.3} & \underline{30.9} \\
    \hline
    \end{tabular}%
  }
  \vspace{-4mm}
  \label{tab:ov-dsgg}
\end{table}

%-------------------------------------------------------------------------
\subsection{Closed-Set DSGG}
\label{sec:results-cs}

Although optimized for open-vocabulary generalization, OvDSGG remains competitive in the closed-set setting (Table~\ref{tab:closedset-dsgg}). It achieves the second-best performance across 11 of 12 \texttt{SGDET} metrics. 
On No-Constraint R@50 it ranks first (53.2), surpassing OED (51.8).
We attribute the remaining gap to OED to Grounding DINO's open-vocabulary architecture. As Liu \etal\ note~\cite{liu2024grounding}, Grounding DINO underperforms its closed-set variant DINO~\cite{zhang2022dino} on closed-set benchmarks. 
Another possible reason is training configuration. We currently freeze the visual backbone and BERT text encoder throughout training, following~\cite{chen2024expanding} to preserve open-vocabulary capabilities. However, in the closed-set setting this could prevent the model from fully adapting its visual--linguistic feature space to Action Genome. OvDSGG's closed-set performance can potentially benefit from a more dedicated optimization strategy, as discussed in Section~\ref{sec:discussion}. 

\begin{table}[htbp]
	\centering
	\caption{Comparison with state-of-the-art DSGG methods on Action Genome for Scene Graph Detection (\texttt{SGDET}), in the closed-set setting. Best and second-best are \textbf{bold} and \underline{underlined}, respectively.}
	\resizebox{0.8\linewidth}{!}{%
		\begin{tabular}{l|cccccc|cccccc}
			\hline
			\multirow{2}{*}{\texttt{SGDET}, IoU\,$\ge 0.5$} & \multicolumn{6}{c|}{R@$K$ $\uparrow$} & \multicolumn{6}{c}{mR@$K$ $\uparrow$} \\
			% \cline{2-13}
			& \multicolumn{3}{c}{With Constr.} & \multicolumn{3}{c|}{No Constr.} & \multicolumn{3}{c}{With Constr.} & \multicolumn{3}{c}{No Constr.} \\
			% \hline
			Method & 10 & 20 & 50 & 10 & 20 & 50 & 10 & 20 & 50 & 10 & 20 & 50 \\
			\hline
			RelDN \cite{zhang2019graphical} & 9.1 & 9.1 & 9.1 & 13.6 & 23.0 & 36.6 & 3.3 & 3.3 & 3.3 & 7.5 & 18.8 & 33.7 \\
			VCTree \cite{tang2019learning} & 24.4 & 32.6 & 34.7 & 23.9 & 35.3 & 46.8 & - & - & - & - & - & - \\
			TRACE \cite{teng2021target} & 13.9 & 14.5 & 14.5 & 26.5 & 35.6 & 45.3 & 8.2 & 8.2 & 8.2 & 22.8 & 31.3 & 41.8 \\
			GPS-Net \cite{lin2020gps} & 24.7 & 33.1 & 35.1 & 24.4 & 35.7 & 47.3 & - & - & - & - & - & - \\
			STTran \cite{cong2021spatial} & 25.2 & 34.1 & 37.0 & 24.6 & 36.2 & 48.8 & 16.6 & 20.8 & 22.2 & 20.9 & 29.7 & 39.2 \\
			APT \cite{li2022anticipatory} & 26.3 & 36.1 & 38.3 & 25.7 & 37.9 & 50.1 & - & - & - & - & - & - \\
			DSG-DETR \cite{feng2023exploiting} & 30.4 & 34.9 & 36.0 & 32.3 & 40.9 & 48.2 & 18.0 & 21.3 & 22.0 & 23.6 & 30.1 & 36.5 \\
			TEMPURA \cite{nag2023unbiased} & 28.1 & 33.4 & 34.9 & 29.8 & 38.1 & 46.4 & 18.5 & 22.6 & 23.7 & 24.7 & 33.9 & 43.7 \\
			OED \cite{wang2024oedonestageendtoenddynamic} & \textbf{33.5} & \textbf{40.9} & \textbf{48.9} & \textbf{35.3} & \textbf{44.0} & \underline{51.8} & \textbf{20.9} & \textbf{26.9} & \textbf{32.9} & \textbf{26.3} & \textbf{39.5} & \textbf{49.5} \\
			% FDSG \cite{yang2025fdsg} & \textbf{35.3} & \textbf{42.9} & \textbf{49.8} & \textbf{37.2} & \textbf{47.2} & \textbf{56.5} & \textbf{22.2} & \textbf{27.8} & \textbf{33.0} & \textbf{27.8} & \textbf{42.0} & \textbf{54.1} \\
			\hline
			OvDSGG (temporal) & \underline{31.3} & \underline{38.2} & \underline{44.7} & \underline{33.4} & \underline{42.4} & \textbf{53.2} & \underline{20.1} & \underline{25.0} & \underline{29.3} & \underline{25.5} & \underline{35.3} & \underline{47.1} \\
			\hline
		\end{tabular}%
	}
	\label{tab:closedset-dsgg}
    \vspace{-2mm}
\end{table}

%-------------------------------------------------------------------------
\subsection{Ablation Study}
\label{sec:ablation}

\subsubsection{Module Effectiveness}
\label{subsec:abl_module}
We assess the contribution of the Triplet Feature Extraction Module (Trp.) by comparing it against a spatial baseline (Spat.), which makes predictions directly from the Spatial Backbone output queries; and comparing against the full model with the Temporal Backbone (Temp.). Results are reported on the open-vocabulary setting in Table~\ref{tab:ov-ablation}. 
The corresponding closed-set ablation is provided in Supplementary Section~B.1.

Adding the Triplet Feature Extraction Module increases every metric for both constraint settings, with a markedly larger relative gain than in the closed-set ablation. %(supplementary). 
This suggests that our proposed Triplet Feature Extraction Module obtains spatial context, providing additional value for classifying zero-shot concepts. 
Adding the temporal module further improves R@$K$ and most mR@$K$ metrics, but regresses on No-Constraint mR@50 and 5 of 6 zR@$K$ scores. As discussed in Section~\ref{sec:settings}, early versions of the open-vocabulary temporal module showed regressions across all metrics, consistent with the catastrophic forgetting reported for OvSGTR~\cite{chen2025datamodelingfullyopenvocabulary}. 
We freeze Grounding DINO and the spatial module during temporal training, which substantially recovers performance. 
The remaining regressions are likely the result of limited recalibration of rare and zero-shot triplet rankings under this frozen regime; see Section~\ref{subsec:abl_vla} for analysis.

\begin{table}[htbp]
	\centering
	\caption{Ablation on Action Genome \texttt{SGDET}, open-vocabulary setting, assessing the contributions of the Spatial (Spat.), Triplet (Trp.) and Temporal (Temp.) modules.}
	\resizebox{\linewidth}{!}{%
		\begin{tabular}{ccc|cccccc|cccccc|cccccc}
			\hline
			\multicolumn{3}{c|}{\multirow{2}{*}{\texttt{SGDET}, IoU\,$\ge 0.5$}} & \multicolumn{6}{c|}{R@$K$ $\uparrow$} & \multicolumn{6}{c|}{mR@$K$ $\uparrow$} & \multicolumn{6}{c}{zR@$K$ $\uparrow$} \\
			% \cline{4-21}
			\multicolumn{3}{c|}{} & \multicolumn{3}{c}{With Constr.} & \multicolumn{3}{c|}{No Constr.} & \multicolumn{3}{c}{With Constr.} & \multicolumn{3}{c|}{No Constr.} & \multicolumn{3}{c}{With Constr.} & \multicolumn{3}{c}{No Constr.} \\
			% \hline
			Spat. & Trp. & Temp. & 10 & 20 & 50 & 10 & 20 & 50 & 10 & 20 & 50 & 10 & 20 & 50 & 10 & 20 & 50 & 10 & 20 & 50 \\
			\hline
			\checkmark &  &  & 22.3 & 28.0 & 35.7 & 22.3 & 29.1 & 38.1 & 10.0 & 12.4 & 15.7 & 12.5 & 17.6 & 25.9 & 9.0 & 12.3 & 17.9 & 10.2 & 15.0 & 22.6 \\
			\checkmark & \checkmark &  & 32.0 & 37.6 & 43.6 & 32.9 & 40.5 & 50.0 & 14.1 & 16.6 & 19.1 & 19.0 & 25.7 & \textbf{36.5} & \textbf{15.5} & \textbf{20.4} & 25.6 & \textbf{17.7} & \textbf{24.6} & \textbf{33.1} \\
			\checkmark & \checkmark & \checkmark & \textbf{33.3} & \textbf{40.1} & \textbf{48.1} & \textbf{33.9} & \textbf{41.5} & \textbf{51.0} & \textbf{16.5} & \textbf{19.8} & \textbf{24.0} & \textbf{19.9} & \textbf{25.8} & 34.8 & 13.6 & 18.8 & \textbf{26.6} & 15.1 & 21.3 & 30.9 \\
			\hline
		\end{tabular}%
	}
    \vspace{-4mm}
	\label{tab:ov-ablation}
\end{table}

\subsubsection{Learnable Visual--Language Alignment}
\label{subsec:abl_vla}

We next isolate the contribution of the learnable text-initialized classification heads of Section~\ref{sec:align_module}. The OvDSGG-frozen variant freezes the head weight matrices $\mathbf{W}^{\text{obj}}$ and $\mathbf{W}^{k}$ at their BERT-initialized values, leaving only the per-class biases trainable as learnable thresholds. Architecturally this mirrors OvSGTR's parameter-free classifier, where text embeddings serving as both semantic prior and decision boundary. However, OvDSGG-frozen does not use OvSGTR's knowledge-distillation retention loss. Therefore, comparing OvDSGG-frozen to OvSGTR isolates the role of distillation, while comparing to the full OvDSGG isolates the learnable head.

\begin{table}[htbp]
	\centering
	\caption{Learnable vs.\ frozen classification heads for visual--language alignment, on Action Genome \texttt{SGDET} in the open-vocabulary setting. Best and second-best are \textbf{bold} and \underline{underlined}, respectively.}
	\resizebox{\linewidth}{!}{%
		\begin{tabular}{l|cccccc|cccccc|cccccc}
			\hline
			\multirow{2}{*}{\texttt{SGDET}, IoU\,$\ge 0.5$} & \multicolumn{6}{c|}{R@$K$ $\uparrow$} & \multicolumn{6}{c|}{mR@$K$ $\uparrow$} & \multicolumn{6}{c}{zR@$K$ $\uparrow$} \\
			% \cline{2-19}
			& \multicolumn{3}{c}{With Constr.} & \multicolumn{3}{c|}{No Constr.} & \multicolumn{3}{c}{With Constr.} & \multicolumn{3}{c|}{No Constr.} & \multicolumn{3}{c}{With Constr.} & \multicolumn{3}{c}{No Constr.} \\
			% \hline
			Method & 10 & 20 & 50 & 10 & 20 & 50 & 10 & 20 & 50 & 10 & 20 & 50 & 10 & 20 & 50 & 10 & 20 & 50 \\
			\hline
			OvSGTR \cite{chen2025datamodelingfullyopenvocabulary} & \underline{12.9} & \underline{16.4} & \underline{20.8} & \underline{12.2} & \underline{16.8} & \underline{23.1} & \underline{7.9} & \underline{9.8} & \underline{11.9} & \underline{9.8} & \underline{14.2} & \underline{20.0} & \underline{3.6} & \underline{4.8} & \underline{6.2} & \underline{3.4} & \underline{5.1} & \underline{8.7} \\
			OvDSGG--frozen & 3.6 & 5.2 & 7.8 & 3.6 & 5.3 & 7.9 & 1.0 & 1.5 & 2.3 & 1.0 & 1.5 & 2.3 & 3.1 & 3.8 & 4.7 & 3.1 & 3.8 & 4.7 \\
			\hline
			OvDSGG (temporal) & \textbf{33.3} & \textbf{40.1} & \textbf{48.1} & \textbf{33.9} & \textbf{41.5} & \textbf{51.0} & \textbf{16.5} & \textbf{19.8} & \textbf{24.0} & \textbf{19.9} & \textbf{25.8} & \textbf{34.8} & \textbf{13.6} & \textbf{18.8} & \textbf{26.6} & \textbf{15.1} & \textbf{21.3} & \textbf{30.9} \\
			\hline
		\end{tabular}%
	}
	\label{tab:abl_vla}
    \vspace{-4mm}
\end{table}

\begin{figure}[htbp]
    \centering
    \includegraphics[width=\linewidth]{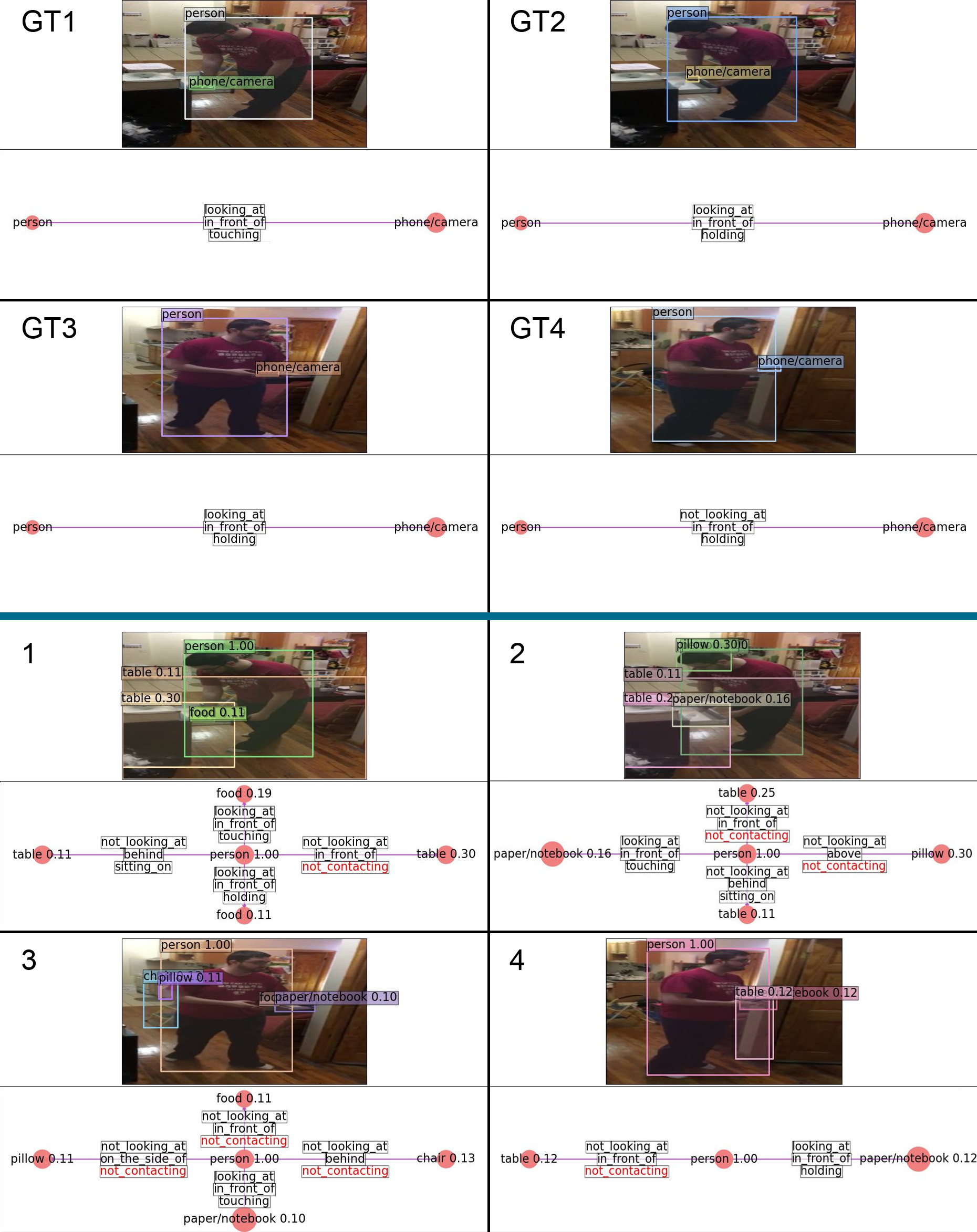}
    \caption{Qualitative results for OvDSGG in the open-vocabulary setting on selected Action Genome frames. Base classes are black; novel classes are red. Ground-truth frames are labeled GT1--GT4; corresponding predictions are labeled 1--4 chronologically. Although not present in the GT scene graph, the model detects novel predicates that are reasonable.}
    \label{fig:qualitative_ov}
\end{figure}

Table~\ref{tab:abl_vla} reports the comparison. Even without distillation, OvDSGG-frozen reaches zR@$K$ broadly comparable to OvSGTR, confirming that the open-vocabulary alignment is preserved by the frozen text prototypes, independently of an explicit retention objective. In contrast, base-class metrics (R@$K$, mR@$K$) fall well below OvSGTR. This demonstrates that the distillation stabilizes the visual feature space during training, and without it the visual side drifts away from the frozen classifier rows, and base-class recognition degrades. Switching to learnable heads (full OvDSGG) dramatically improves every metric: R@$K$ and mR@$K$ recover and substantially surpass OvSGTR, and zR@$K$ also rises sharply. This is because the novel-class rows still receive no gradient by using the masked loss (Section~\ref{sec:align_module}) and thus retain their text-embedding values. The two comparisons jointly validate the decoupled design: structural retention provides open-vocabulary alignment without distillation, while per-class learnable capacity adapts base-class boundaries to Action Genome.

%-------------------------------------------------------------------------
\subsection{Qualitative Results}
\label{sec:qualitative}

Figure~\ref{fig:qualitative_ov} shows open-vocabulary OvDSGG predictions on an Action Genome clip in which a man examines a phone and walks through a doorway. Base classes are colored black; novel classes are colored red. OvDSGG correctly recovers most seen relations and produces reasonable predictions for unseen concepts, \eg the novel predicate \texttt{not\_contacting}. Notably, many objects and predicates are not labelled in the GT annotations, which is a known issue with Action Genome~\cite{cong2021spatial,wang2024oedonestageendtoenddynamic}, while OvDSGG can still predict them. 

%-------------------------------------------------------------------------
\section{Discussion and Limitations}
\label{sec:discussion}

OvDSGG's principal limitation is the residual regression of the open-vocabulary temporal module on No-Constraint mR@50 and most zR@$K$ metrics. Freezing the Grounding DINO detector and spatial components during temporal training mitigates the catastrophic forgetting observed in earlier runs, but the frozen regime limits the model's ability to recalibrate rare- and zero-shot triplet rankings using temporally enriched features. This could potentially be resolved through temporal modules that operate on the language-aligned rather than the visual side of the representation, or through parameter-efficient temporal adapters that leave the aligned feature space intact.  

A second direction follows from the qualitative analysis (Section~\ref{sec:qualitative}): visually similar common Base classes occasionally dominate rare Base and Novel alternatives. This is the long-tailed bias inherited from Action Genome, also reflected in the gap between R@$K$ and mR@$K$. Although open-vocabulary alignment alone mitigates this bias for Novel categories, OvDSGG could plausibly benefit further from explicit debiasing strategies such as the causal-intervention approach of Tang \etal~\cite{tang2020unbiased}, applied on top of the visual--language alignment.

%-------------------------------------------------------------------------
\section{Conclusion}
\label{sec:conclusion}

We introduced OvDSGG, the first end-to-end framework for open-vocabulary dynamic scene graph generation, alongside a rigorous open-vocabulary DSGG benchmark adapted from Action Genome. OvDSGG bridges an open-vocabulary detector and an OED-style temporal module via two novel components: a Triplet Feature Extraction Module that supplies the per-query triplet representation, and a Visual--Language Alignment Module whose learnable text-initialized classification heads preserve novel-class alignment structurally, without an explicit distillation loss. On the proposed benchmark, OvDSGG outperforms all tested baselines on every open-vocabulary metric, with zR@$K$ scores 10.0--20.4\,pp higher than the next-best baseline, and remains competitive with state-of-the-art closed-set DSGG models. These results establish OvDSGG as a foundation for future open-vocabulary DSGG research.

%-------------------------------------------------------------------------
\section*{Acknowledgements}
This work has been supported by the 
Centre for Spatial Intelligence (RCSI) at University
of Bath,  the European Union under grant agreement no. 101136006-XTREME,
the European Innovation Council under grant agreement no. 1012575-36-CEREBRIS,
the MWK of Lower Saxony within Hybrint (VWZN4219) and LCIS (VWZN4704), the DFG under Germany’s Excellence Strategy within the Cluster of Excellence PhoenixD (EXC2122) and Quantum Frontiers (EXC2123).

\bibliographystyle{splncs04}
\bibliography{references}

\newpage
% ============================================================
\section*{\centering OvDSGG - Supplementary Materials}

\setcounter{section}{0}
\renewcommand{\thesection}{\Alph{section}}
% ---------------------------------------------------------------
\setcounter{figure}{0}
\setcounter{table}{0}
\renewcommand{\thefigure}{A\arabic{figure}}
\renewcommand{\thetable}{A\arabic{table}}
%-------------------------------------------------------------------------
\section{Statistics of Base/Novel Split on Action Genome}
\label{sec:stats}

\begin{figure}[htbp]
    \centering
    \includegraphics[width=0.9\linewidth]{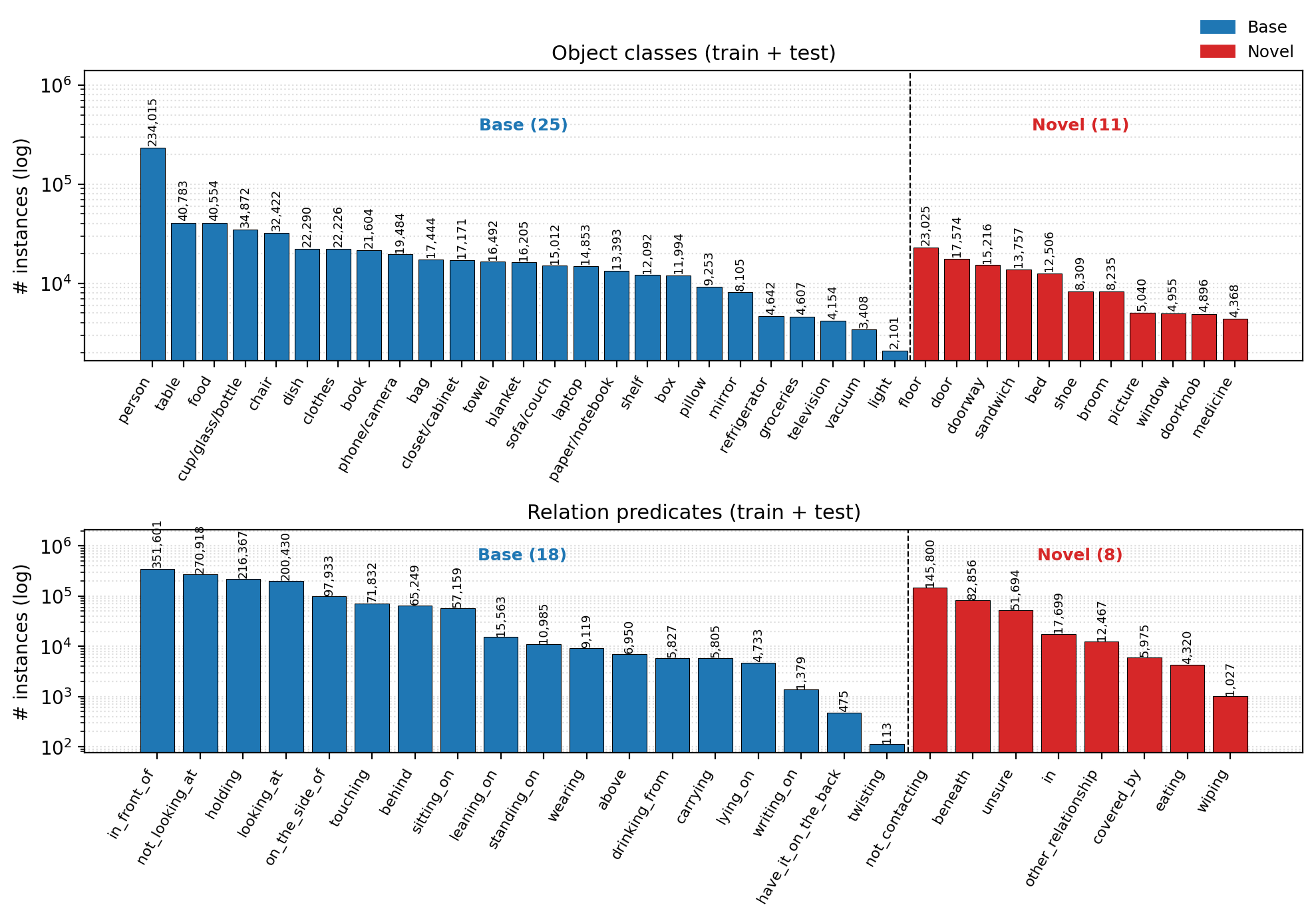}
    \caption{Object and predicate class distribution on our Base/Novel split of Action Genome dataset. }
    \label{fig:ag_base_novel_stats}
    \vspace{-2mm}
\end{figure}

\noindent Figure~\ref{fig:ag_base_novel_stats} summarizes the per-class instance statistics of the Action Genome dataset under our Base/Novel partition, aggregated over the train and test splits. 
Note that the data sample statistics are different from the $70\%-30\%$ partition on the category vocabularies, which we follow OvSGTR~\cite{chen2024expanding}. 
The label set comprises $36$ object categories ($25$ Base, $11$ Novel) and $26$ relation predicates ($18$ Base, $8$ Novel). 
Base categories contain $639{,}176$ object instances ($84.4\%$) and $1{,}392{,}438$ relation occurrences ($81.2\%$), whereas the Novel partition contributes $117{,}881$ object instances ($15.6\%$) and $321{,}838$ relation occurrences ($18.8\%$). 

Importantly, the Novel partition is not simply the rare tail of the distribution. Several Novel categories are in fact among the most
frequent in the dataset, \eg, \texttt{floor} is the sixth most common object overall (outranking $19$ Base classes), and the Novel predicate \texttt{not\_contacting} is the fifth most common relation. At the same time, the Novel partition does contain genuinely scarce categories, such as \texttt{wiping} ($1{,}027$ occurrences) among predicates and \texttt{medicine}, \texttt{doorknob}, and \texttt{window} ($\leq\!5{,}000$ instances) among objects. We use this mixture of head and tail concepts, so that it ensures that the open-vocabulary evaluation probes a model's ability to generalize to semantically novel categories rather than merely to low-shot ones, while still preserving enough Novel-class supervision in the test split to yield statistically meaningful per-class metrics.

%-------------------------------------------------------------------------
\section{Additional Results on Closed-Vocabulary Setting}
\label{sec:eval_close}

\subsection{Module Effectiveness Ablation}
\label{sec:abl_close}

Table~\ref{tab:cs-ablation} reports the closed-set counterpart to the open-vocabulary module-effectiveness ablation presented in the main paper. Adding the spatial module to the baseline improves every R@$K$ and mR@$K$ metric across both constraint settings, indicating that OvDSGG's pair-wise interaction decoder effectively integrates relevant within-frame context. The subsequent addition of the temporal module yields further consistent gains across all metrics, confirming that temporal context aggregation is also valuable in the closed-set regime. These relative gains mirror those reported in OED's own ablation~\cite{wang2024oedonestageendtoenddynamic}, indicating that the interaction modules retain their effectiveness when ported into OvDSGG's open-vocabulary architecture. The spatial-module gains are, however, noticeably smaller in relative terms than in the open-vocabulary setting, supporting the observation in the main paper that spatial context contributes more strongly when novel categories must be inferred without direct supervision.

\begin{table}[htbp]
	\centering
	\caption{Module-effectiveness ablation on Action Genome \texttt{SGDET} in the \emph{closed-set} setting, assessing the contributions of the Spatial (Spat.), Triplet (Trp.) and Temporal (Temp.) modules. Counterpart to Table~3 of the main paper.}
	\resizebox{0.8\linewidth}{!}{%
		\begin{tabular}{ccc|cccccc|cccccc}
			\hline
			\multicolumn{3}{c|}{\multirow{2}{*}{\texttt{SGDET}, IoU\,$\ge 0.5$}} & \multicolumn{6}{c|}{R@$K$ $\uparrow$} & \multicolumn{6}{c}{mR@$K$ $\uparrow$} \\
			% \cline{4-15}
			\multicolumn{3}{c|}{} & \multicolumn{3}{c}{With Constr.} & \multicolumn{3}{c|}{No Constr.} & \multicolumn{3}{c}{With Constr.} & \multicolumn{3}{c}{No Constr.} \\
			% \hline
			Spat. & Trp. & Temp. & 10 & 20 & 50 & 10 & 20 & 50 & 10 & 20 & 50 & 10 & 20 & 50 \\
			\hline
			\checkmark &  &  & 28.9 & 35.8 & 42.9 & 30.6 & 39.3 & 50.0 & 16.9 & 22.0 & 27.0 & 21.7 & 31.8 & 43.8 \\
			\checkmark & \checkmark &  & 29.5 & 36.4 & 43.3 & 31.6 & 40.8 & 52.0 & 19.5 & 24.2 & 28.3 & 24.9 & 33.6 & 44.9 \\
			\checkmark & \checkmark & \checkmark & \textbf{31.3} & \textbf{38.2} & \textbf{44.7} & \textbf{33.4} & \textbf{42.4} & \textbf{53.2} & \textbf{20.1} & \textbf{25.0} & \textbf{29.3} & \textbf{25.5} & \textbf{35.3} & \textbf{47.1} \\
			\hline
		\end{tabular}%
	}
    \vspace{-4mm}
	\label{tab:cs-ablation}
\end{table}

\subsection{Additional Qualitative Results}
\label{sec:qualitative_close}

\begin{figure}[htbp]
    \centering
    \includegraphics[width=\linewidth]{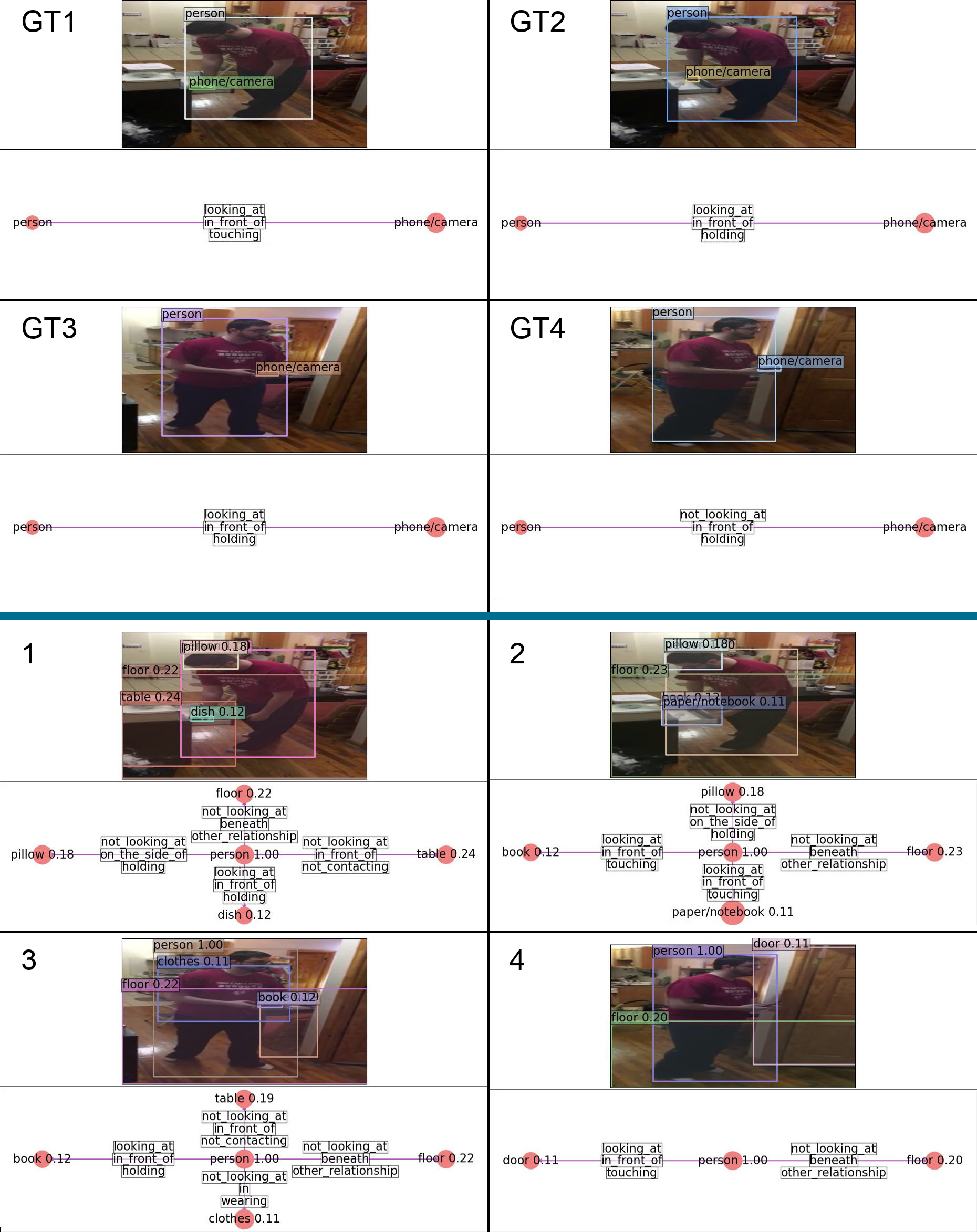}
    \caption{Qualitative results for OvDSGG in the closed-set setting on selected Action Genome frames. Ground-truth frames are labeled GT1--GT4; corresponding predictions are labeled 1--4 chronologically. Although not present in the GT scene graph, the model detects predicates that are plausible. Compare with the open-vocabulary results in the main paper.}
    \label{fig:qualitative_close}
\end{figure}

Figure~\ref{fig:qualitative_close} shows closed-set OvDSGG predictions on the same Action Genome clip used in the main paper's open-vocabulary qualitative results (a man examining a phone and walking through a doorway). In the closed-set setting the model has seen all categories during training. OvDSGG produces a reasonable scene graph for the clip, correctly predicting objects such as \texttt{floor}, \texttt{table}, \texttt{door}, and \texttt{clothes}, although the rare \texttt{phone/camera} class remains difficult.

\end{document}